\documentclass{article}

\usepackage[final]{corl_2022} %
\usepackage{wrapfig,lipsum,booktabs}
\usepackage{amsmath}
\usepackage{amssymb}
\usepackage{amsfonts}
\usepackage{graphicx}
\usepackage{color}
\usepackage{subcaption}
\usepackage[capitalise]{cleveref}
\usepackage{enumitem}

\title{PI-QT-Opt: Predictive Information Improves Multi-Task Robotic Reinforcement Learning at Scale}

\author{Kuang-Huei Lee\thanks{Equal contribution} \qquad Ted Xiao \qquad Adrian Li \qquad Paul Wohlhart \qquad Ian Fischer \qquad Yao Lu$^*$ \\
Google Research \\
\texttt{[leekh, tedxiao, alhli, wohlhart, iansf, yaolug]@google.com}
}

\newcommand{\taskone}{Experiment 1}
\newcommand{\tasktwo}{Experiment 2}
\newcommand{\taskthree}{Experiment 3}
\newcommand{\taskfour}{Experiment 4}
\newcommand{\taskfive}{Experiment 5}
\newcommand{\tasksix}{Experiment 6}

\begin{document}
\maketitle

\begin{abstract}
The \emph{predictive information}, the mutual information between the past and future, has been shown to be a useful representation learning auxiliary loss for training reinforcement learning agents, as the ability to model what will happen next is critical to success on many control tasks.
While existing studies are largely restricted to training specialist agents on single-task settings in simulation, in this work, we study modeling the predictive information for robotic agents and its importance for general-purpose agents that are trained to master a large repertoire of diverse skills from large amounts of data.
Specifically, we introduce Predictive Information QT-Opt (PI-QT-Opt), a QT-Opt agent augmented with an auxiliary loss that learns representations of the predictive information to solve up to 297 vision-based robot manipulation tasks in simulation and the real world with a single set of parameters.
We demonstrate that modeling the predictive information significantly improves success rates on the training tasks and leads to better zero-shot transfer to unseen novel tasks.
Finally, we evaluate PI-QT-Opt on real robots, achieving substantial and consistent improvement over QT-Opt in multiple experimental settings of varying environments, skills, and multi-task configurations.
\end{abstract}

\keywords{deep reinforcement learning, robot manipulation, multi-task learning} 
\section{Introduction}
\label{sec:intro}

Real robotic control systems are often partially observable and non-Markovian, and include high-dimensional observations, such as pixels.
In such systems, we can learn representations by explicitly modeling the mutual information between consecutive states and actions -- the \emph{predictive information}~\citep{bialek1999predictive} -- to facilitate policy and value learning~\citep{lee2020predictive,oord2018representation}.
When learning a predictive information representation between a state, action pair and its subsequent state, the learning task is equivalent to modeling environment dynamics~\citep{anand2019unsupervised}.
In this work, we are interested in training multi-task generalist agents~\citep{reed2022generalist,lee2022multi,saycan2022arxiv} that can master a wide range of robotics skills in both simulated and real environments by learning from a large amount of diverse experience.
We hypothesize that modeling the predictive information will give latent representations that capture environment dynamics across multiple tasks, making it simpler and more efficient to learn a generalist policy.
We also hypothesize that such a generalist agent may do a better job of transferring to real-world environments and novel tasks unseen during training.
While existing studies of learning predictive information representations in RL~\citep{lee2020predictive,oord2018representation,anand2019unsupervised,stooke2021decoupling,chen2022empirical} have largely been limited to learning single-task specialist agents in simulated environments such as DM-Control~\citep{tassa2018dmcontrol} and Atari~\citep{bellemare13arcade}, our hypotheses can be seen as extending the generalization results in \citet{lee2020predictive}, which showed both more sample-efficient learning and better fine-tuning on unseen tasks.
We investigate both of these hypotheses in this work.

We combine a predictive information auxiliary loss with QT-Opt~\citep{kalashnikov2018scalable}, a model-free off-policy reinforcement learning method that been shown to work well on vision-based continuous control problems.
QT-Opt is able to leverage large-scale, multi-task datasets in simulation and the real world~\citep{kalashnikov2018scalable,mtopt2021arxiv,xiao2020nonblocking,awopt2021corl}.
We train a task-conditioned QT-Opt agent~\citep{kalashnikov2018scalable} with a predictive information auxiliary loss similar to~\citet{lee2020predictive}, which we refer to as Predictive Information QT-Opt (PI-QT-Opt). 
We study various simulated and real robot environments using an Everyday Robots manipulator arm~\citep{edrwebsite}, including a large diverse set of real-world environments with up to 297 challenging vision-based manipulation tasks in a kitchen setting~\citep{saycan2022arxiv} (See \Cref{subsec:context_conditioning} for task definitions).

\citet{lee2020predictive} showed the benefits of predictive information regularization for accerlerating policy learning and quickly finetuning a policy trained on one task to solve a related task in the same environment.
Our experiments show that predictive information regularization additionally gives substantial benefits in two challenging zero-shot settings: from simulation to real-world robots, and from one set of training objects to novel objects in both simulation and real environments.
We demonstrate that PI-QT-Opt significantly outperforms QT-Opt in terms of success rate on training tasks in simulation.
When evaluated on tasks that are unseen during training, modeling the predictive information increases the zero-shot success rate substantially. 
We verify these improvements on real robots via sim-to-real transfer, and observe that PI-QT-Opt significantly outperforms QT-Opt.  %

Our primary contributions are:
\begin{itemize}[leftmargin=*,topsep=0pt]
\item We validate that modeling the predictive information is an effective auxiliary task for learning multi-task generalist robot control agents.
\item We show that simple forms of task conditioning are sufficient to allow QT-Opt learn to solve large numbers of tasks simultaneously, avoiding some of the complexity of earlier multi-task approaches to QT-Opt~\citep{mtopt2021arxiv}.
\item We verify the improvements through large-scale real-world experiments, including training a reward-based agent that performs well on 297 real robotic control tasks.
\item We show that the predictive information helps zero-shot generalization to unseen tasks.
\item We demonstrate that PI-QT-Opt can train only on simulated environments and transfer to real robots more effectively than our QT-Opt baseline.
\end{itemize}

\section{Related Work}
\label{sec:related_work}

\paragraph{Predictive Information Representations.}
Previous studies~\cite{lee2022piars,lee2020predictive,anand2019unsupervised,oord2018representation,chen2022empirical,rakelly2021mutual,stooke2021decoupling,yan2020learning} have shown that predictive information~\citep{bialek1999predictive} is an effective auxiliary or representation learning objective for RL agents or planning.
This result can be connected to findings in neuroscience that suggested that the brain maximizes predictive information at an abstract level~\citep{friston2005theory,rao1999predictive}.
Broadly, our work differs from those approaches by focusing on multi-task, vision-based robot learning in both simulated and real-world environments.
This enables us to verify improvements and study broader generalization properties in realistic settings.
Additionally, with the exception of \citep{lee2020predictive}, most such works do not explicitly learn a compressed representation of the predictive information.
Finally, because the predictive information can model the underlying environment dynamics, which are stationary, its use as an auxiliary loss can also be seen as a representation regularizer for RL agents, possibly helping avoid overfitting to any specific value function during training~\citep{dabney2021value}.

\paragraph{Robot Manipulation.}
We focus on two main categories of related work in robot manipulation: multitask robot learning methods and real world data-driven robotics methods.
One family of approaches for multitask robot learning focuses on supervised learning of multitask control policies to maximize few-shot performance via meta-learning~\citep{finn2017visualimitation,rahmatizadeh2017multitask,james2018fewshot} or direct zero-shot performance via behavioral cloning~\citep{fox2019multi,jang2021bc,robomimic2021,shridhar2021cliport}. 
However, methods based on imitation learning require expensive expert demonstrations and struggle to improve autonomously with on-policy learning.
On the other hand, reinforcement learning (RL) based approaches are able to bootstrap without a large amount of expert data and are able to continuously improve from their own experience. 
While some RL methods combine subtasks~\citep{barto2004intrinsically,dietterich2000hierarchical} or map tasks to individual policies~\citep{da2012learning,Deisenroth2014MultitaskPS}, we wish to learn a single shared policy.
Recent work focuses on learning a single multi-task policy with both on-policy~\citep{yu2020meta} and off-policy RL methods~\citep{yu2020gradient,yang2020multitask,sodhani2021multitask}, and have shown promising results on a variety of robot manipulation tasks in simulation~\citep{yu2020meta}.
Some approaches to real-world robot learning for manipulation utilize RL with real robot data collection or simulation with domain randomization ~\citep{awopt2021corl,shridhar2021cliport,kalashnikov2018scalable,mtopt2021arxiv,tobin2017domain,chebotar2018sim,peng2018sim,akkaya2019solving,james2019sim,zhu2020ingredients}, while other approaches focus on imitation learning from expert demonstrations~\citep{jang2021bc,lopes2007affordance,sharma2019third,ratliff2007imitation,fang2019survey,khansari2020action}.
MT-Opt~\citep{mtopt2021arxiv} is closely related to our approach; it also extends QT-Opt~\citep{kalashnikov2018scalable} to a real world multitask robot learning setting.
While MT-Opt suggests that data routing and data sharing is very important, it is quite challenging to scale this approach to the hundreds of tasks that we consider.
Thus, we take a different approach from MT-Opt, focusing on learning better representations for a single task-conditioned critic that does not use any data routing.

\section{Methods}
\label{sec:methods}

\begin{figure}[t]
    \centering
    \includegraphics[width=1.0\linewidth]{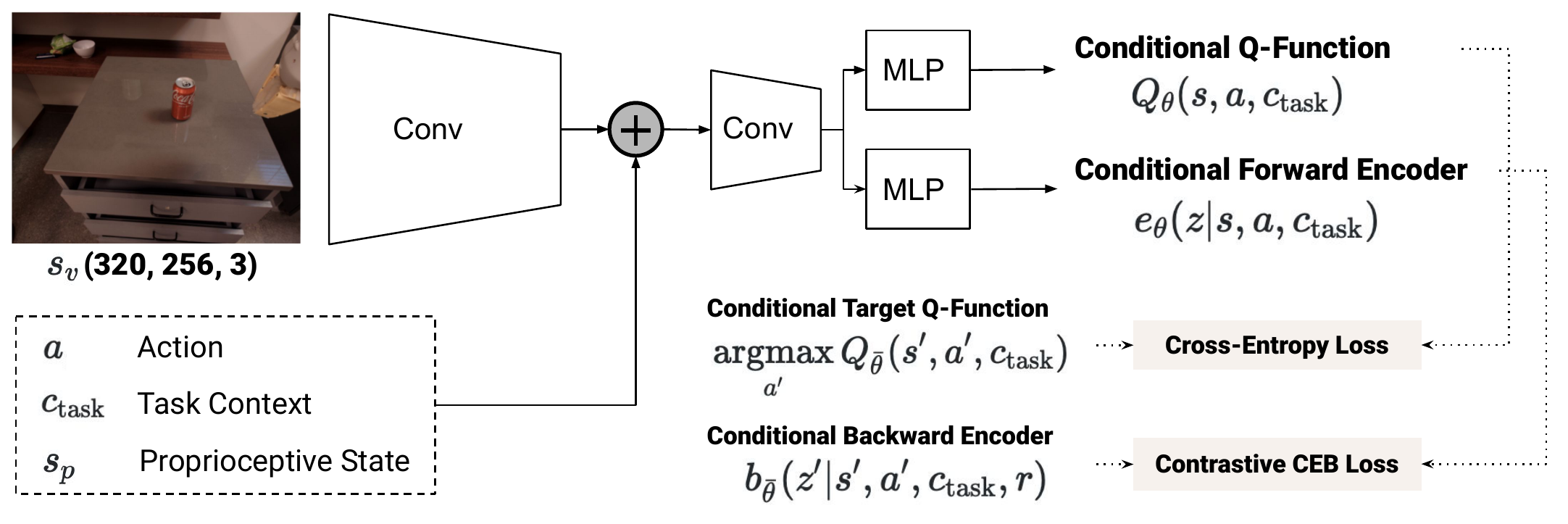}
    \caption{
        An overview of the PI-QT-Opt system for multi-task robotic reinforcement learning.
        The contrastive CEB loss is an auxiliary objective to Q-learning.
        $s = (s_v, s_p)$.
        See \Cref{sec:methods} for details.
    }
    \label{fig:piqtopt_diagram}
\end{figure}

We describe the details of Predictive Information QT-Opt (PI-QT-Opt) in \Cref{subsec:piqtopt}, and task context conditioning for enabling multi-task robot RL at scale in \Cref{subsec:context_conditioning}.
More implementation details are described in \Cref{sec:impl_details}.
\Cref{fig:piqtopt_diagram} presents an overview of the our system.

\subsection{Predictive Information QT-Opt}
\label{subsec:piqtopt}

\paragraph{Predictive Information.}
The \emph{Predictive Information}~\cite{bialek1999predictive} is the mutual information between the past and the future, $I(\textit{past};\textit{future})$.
It has been shown that the predictive information is an effective auxiliary loss for RL agents~\citep{lee2020predictive,oord2018representation}.
From here on, we will denote the past by $X$ and the future by $Y$.
\citet{lee2020predictive} argues that a learned representation $Z$ of the predictive information should be compressed with respect to $X$, based on the observation in \citet{bialek1999predictive} that $H(X)$, the entropy of the past, grows more quickly than $I(X;Y)$.
Following \citep{lee2020predictive}, we use the Conditional Entropy Bottleneck (CEB) \citep{fischer2020conditional} to learn the representation $Z$, utilizing the same variational bound on CEB:
\begin{align}
CEB &\equiv \min_Z \beta I(X;Z|Y) - I(Y;Z) \\
&\leq \min_Z \mathrm{E}_{x,y,z \sim p(x,y)e(z|x)} \beta \log \frac{e(z|x)}{b(z|y)} - \log \frac{b(z|y)}{\frac 1 K \sum_{k=1}^K b(z|y_k)} \label{eq:ceb}
\end{align}
where $(x,y)$ are sampled from the data distribution, $p(x,y)$, $e(z|x)$ is the learned \emph{forward encoder} distribution, $b(z|y)$ is the learned variational \emph{backward encoder} distribution, $\beta$ is a Lagrange multiplier that controls how strongly compressed the learned representation $Z$ is, with smaller values corresponding to less compression, and $K$ is the number of examples in a mini-batch during training.
The second term of \Cref{eq:ceb} corresponds to the contrastive InfoNCE bound~\citep{oord2018representation,poole2019variational} on mutual information $I(Y;Z)$.
Following the CEB implementation of \citet{lee2021compressive}, we choose $e(z|x)$ and $b(z|y)$ to be parameterized von Mises-Fisher distributions.
Details are described in \Cref{subsec:pi_aux_details}.

\paragraph{QT-Opt.}
QT-Opt is an offline actor-critic algorithm where only the critic is explicitly learned.
It learns the Q-function (or critic) by minimizing Bellman errors:
\begin{align}
\mathcal{E}(\theta) = \mathbb{E}_{(s,a,s') \sim p(s,a,s')} D \left[ Q_\theta(s,a), Q_T(s,a,s') \right] \label{eq:bellman_loss}
\end{align}
where $\theta$ is the set of model parameters, $s$ and $s'$ are state observations, $a$ is the action taken, $D$ is some divergence (QT-Opt uses cross-entropy), $Q_\theta$ is the learned state-action value function, and $Q_T = r(s,a) + \gamma V(s')$ gives a \emph{target} value for the given transition $(s,a,s')$.
$V(s') = \min_{i \in [1,2]} Q_{\bar{\theta}_i} (s', \pi_{\bar{\theta}_1}(s'))$ is a Double DQN state value function~\citep{hasselt2010double,van2016deep,fujimoto2018addressing}, which for QT-Opt is computed by using two lagged versions of the parameters $\theta$, $\bar{\theta}_1$ and $\bar{\theta}_2$, with different lagging methods.
The QT-Opt policy, $\pi_{\bar{\theta}_1}(s) = \arg \max_{a} Q_{\bar{\theta}_1}(s, a)$, is optimized directly at each environment step using the cross-entropy method (CEM)~\citep{rubinstein2004cross}.
In CEM, $N$ actions are sampled from a Guassian over the action space, the best $M < N$ actions as measured by $Q_{\bar{\theta}_1}$ are used to estimate the mean and variance of a new Guassian, from which another $N$ samples are drawn.
This is repeated a fixed number of steps, converging towards a narrow Gaussian over the part of the action space that the critic believes will perform best at the current state.
See \citep{kalashnikov2018scalable} for further details.

\paragraph{PI-QT-Opt.}
As shown in \Cref{fig:piqtopt_diagram}, PI-QT-Opt combines a predictive information auxiliary similar to that introduced in \citet{lee2020predictive} with the QT-Opt architecture.
We define the past ($X$) to be the current state and action, $(s, a)$, and the future ($Y$) to be the next state, next optimal action, and reward, $(s', a', r)$.
A state $s$ includes an RGB image observation and proprioceptive information.
Image observations are processed by a simple conv net, the output of which is mixed with action, proprioceptive state, and the current task context (described in \Cref{subsec:context_conditioning}) using additive conditioning.
A second simple conv net processes the combined state representation.
All of the convolutional parameters are shared by both the forward encoder $e_{\theta}$ for modeling the predictive information (as in \Cref{eq:ceb}) and the Q-function $Q_{\theta}$ (as in \Cref{eq:bellman_loss}), but the shared representation output from this is further processed by separate MLPs, to allow each loss to specialize its representation as needed, while still allowing the predictive information loss to influence the shared convolutional representation.
Not shown in \Cref{fig:piqtopt_diagram} is that the target Q-function $Q_{\bar{\theta}_1}$ and backward encoder $b_{\bar{\theta}_1}$ for modeling the predictive information also share the same base lagged and non-trainable convolutional representation, but the backward encoder has its own trainable MLP, in order to learn any differences in dynamics when trying to predict the past from the future, rather than predicting the future from the past, as the forward encoder does.
In addition, we concatenate the convolutional representation with observed reward $r(s, a)$ as the input to the backward encoder MLP head.

We find that adding a predictive information auxiliary loss is an easy way to give substantial performance improvements to our chosen RL algorithm, as in \citet{lee2020predictive} which introduced Predictive Information Soft Actor-Critic (PI-SAC). 
However, we note that PI-SAC on its own was unable to solve our tasks, yielding close-to-zero success rates, and neither was SAC~\citep{haarnoja2018soft}\footnote{With our best effort, we were unable to get SAC and PI-SAC working on our tasks (See \Cref{sec:qtopt_v_sac})}, which may indicate that the choice of base RL algorithm is still critical.

\subsection{Multi-Task Context Conditioning}
\label{subsec:context_conditioning}

In order to learn one general-purpose agent for multiple tasks, we condition the Q-functions and the Predictive Information auxiliary on a \emph{task context}, which describes the specific task that we wish the agent to perform, as illustrated in~\Cref{fig:piqtopt_diagram}.
In our setting, a task involves a robot skill and a set of objects that the robot should interact with.
We use two practical implementations of task context in different robot manipulation settings (\Cref{subsec:env_overview}).
One is image-based, where a task is specified with the initial image, the initial object locations, and the skill type as depicted in \Cref{fig:experiments}.
It only considers locations and skill types and thus could enable good generalization across different and even novel objects.
The other one is language-based, where tasks are specified with natural language, similar to~\citet{saycan2022arxiv}.
Details of these two implementations are described in \Cref{sec:task_conditioning_details}.
A generalization of the language-based approach is considered in \Cref{sec:extend_nat_lang}.

An alternative approach to extending QT-Opt for multi-task learning is having one critic head per task with a shared base encoder, as used in MT-Opt~\citep{mtopt2021arxiv}.
However, MT-Opt found that a multi-headed architecture performed worse than a single-headed architecture; in addition, we note that a multi-headed strategy is practically prohibitive to scale to the order of hundreds of tasks.

\section{Experimental Setup}
\label{sec:exp_setup}

To analyze how PI-QT-Opt compares with QT-Opt across different multi-task robotic learning scenarios, we explore a variety of challenging simulation and real vision-based robotic manipulation environments.
While prior results on large-scale robotic grasping focused on a limited set of tasks~\citep{kalashnikov2018scalable,mtopt2021arxiv}, we verify the robustness and scalability of PI-QT-Opt by studying many different environments across hundreds of different tasks in the real world.

\subsection{Robot Manipulation Tasks}
\label{subsec:env_overview}

\begin{figure}[t]
    \centering
    \includegraphics[width=1.0\linewidth]{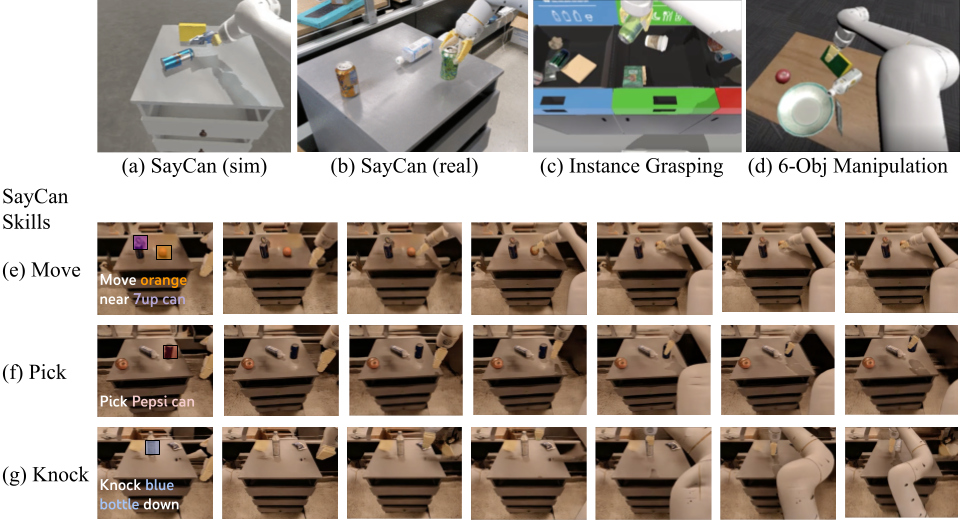}
    \caption{
        Example tasks in simulation and real-world environments using the \href{www.everydayrobots.com}{Everyday Robots manipulator arm}.
        See \Cref{subsec:env_overview} for details. 
        The first images of (e), (f) and (g) show target object locations used in the image-based task context conditioning method we designed. 
        In real-world environments, the object locations are detected using a vision model (\Cref{subsec:real_eval}). 
        An overlay image is created based on the object locations and the skill type, where we use different colors to indicate different skills. 
        This overlay image is only created for the first image of an episode. 
        Then, the first image and the overlay image are used as task context throughout the whole episode.
    }
    \label{fig:experiments}
\end{figure}

We study 6 different multi-task, vision-based robotic manipulation settings in 3 different environments in simulation.
Four of the manipulation settings have the corresponding hardware setup permitting real-world evaluation.

\textit{\taskone{}:} \textbf{(Sim \& Real) SayCan Move skill} (\Cref{fig:experiments}(a),(b),(e)). 
We use the kitchen environment from SayCan~\citep{saycan2022arxiv} which contains 17 objects that spawn on a countertop. 
The simulation environment is shown in \Cref{fig:experiments}(a) and the real countertop is shown in \Cref{fig:experiments}(b). 
The Move skill contains 272 tasks of the form ``move object A near object B''. 
246 tasks are used during training and the 26 remaining tasks are held out.
Image-based task conditioning is used.\\
\textit{\tasktwo{}:} \textbf{(Sim \& Real) SayCan Pick skill} (\Cref{fig:experiments}(a),(b),(f)). 
Same environment as \taskone{}. 
The Pick skill encompasses picking up each of the individual objects, for a total of 17 tasks. 
12 tasks are used during training and the remaining 5 tasks are held out.
Image-based task conditioning is used, as described in~\Cref{subsec:context_conditioning}.\\
\textit{\taskthree{}:} \textbf{(Sim \& Real) SayCan Knock skill} (\Cref{fig:experiments}(a),(b),(g)). 
Same environment as \taskone{}. 
The Knock skill contains 8 tasks testing knocking over a can or bottle. 
7 tasks are used during training and 1 task is held out.
Image-based task conditioning is used.\\
\textit{\taskfour{}:} \textbf{(Sim \& Real) SayCan 297 tasks, All skills} (\Cref{fig:experiments}(a),(b),(e)-(g)). 
Same environment as \taskone{}. 
This task set includes all 3 SayCan skills (297 tasks).
265 tasks are used during training and the 32 remaining tasks are held out.
Image-based task conditioning is used.\\
\textit{\taskfive{}:} \textbf{(Sim Only) Instance Grasping} (\Cref{fig:experiments}(c)). 
A sampled subset of 37 different trash objects are placed randomly in bins. 
A vision model provides a target object for the robot to grasp.
The episode is successful if the robot lifts the target object.
Image-based task conditioning is used.\\
\textit{\tasksix{}:} \textbf{(Sim Only) 6-Object Manipulation} (\Cref{fig:experiments}(d)). 
A fixed set of 6 objects are placed randomly on a table, similar to~\citep{jang2021bc}. 
There are 30 separate tasks comprising the manipulation skills of picking, pushing, and pick and place.
Language-based task conditioning is used.\\   

In all of the experiments, we use an Everday Robots manipulator robot~\citep{edrwebsite} with parallel-jaw grippers, an over-the-shoulder camera, and a 7-DoF arm.
The robot has a proprioceptive observation space that includes the RGB camera image, the arm pose, and the gripper angle.
For the action space, learned policies control the robot via relative position control of the end effector.
In the simulation-only environments, we utilize blocking control, where the policy waits until the previous action completes before planning the next action.
Motivated by faster and more reactive robot motions for real world evaluations, we utilize concurrent control~\citep{xiao2020nonblocking} in all SayCan experiments, which means that the current action is computed while the previous action is still executing.
We provide additional implementation details in the \Cref{sec:impl_details}.

\subsection{Reducing the Sim-to-Real Gap with CycleGAN}
\label{subsec:sim2real}
We train our PI-QT-Opt and QT-Opt models in simulated environments that roughly match the real world evaluation environments.
In order to reduce the simulation-to-real (sim-to-real) gap when deploying policies on real robots, we train a RetinaGAN~\citep{ho2021retinagan} model, a version of CycleGAN~\citep{zhu2017unpaired}, to transform simulated robot images to look more realistic while preserving general object structure, following the sim-to-real setup in \citep{xiao2020nonblocking}.
This enables our method to train purely on RetinaGAN-transformed simulation images and directly transfer to the real world.
We apply this image transformation to all the SayCan models, which we evaluate in both simulation and the real world.

\subsection{Training and Evaluation Protocols}
For each of the experimental settings in \Cref{subsec:env_overview}, we create a sparse binary reward (success or failure) in simulation based on ground truth object poses. 
For each environment, we follow the large scale asynchronous distributed training procedure in~\citep{kalashnikov2018scalable} and train QT-Opt agents and PI-QT-Opt agents with the same hyperparameters with a batch size of 4096 using 16 TPUv2.
In simulation, we evaluate models with 700 episodes and compute their success rates.
For the real-world evaluations, we test each policy on 50 episodes of standardized starting scenarios for a fair comparison between policies. 
More details are described in \Cref{subsec:training_details} and \Cref{subsec:eval_details}.

\section{Experimental Results}
\label{sec:exp_results}

We discuss experimental results on the manipulation tasks introduced in \Cref{subsec:env_overview} in \Cref{subsec:saycan_exp_training} and \Cref{subsec:real_eval}.
We report mean and one standard deviation of success rate over 3 training runs with different random seeds for each model.  %
We analyze the relationship between predictive information and agent performance in \Cref{subsec:pi_analysis} and information compression in \Cref{sec:compression_exp}.

\subsection{Evaluation in Simulated Environments}
\label{subsec:saycan_exp_training}

\begin{figure}[t]
    \centering
    \begin{subfigure}{0.32\textwidth}
        \centering
        \includegraphics[width=1\linewidth]{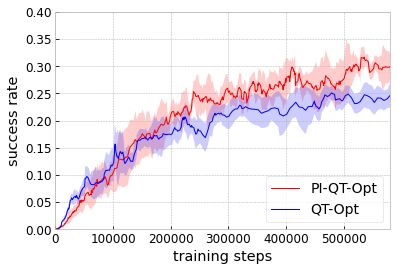}
        \caption{%
            Move Skill
        }
    \end{subfigure}
    \begin{subfigure}{0.32\textwidth}
        \centering
        \includegraphics[width=1\linewidth]{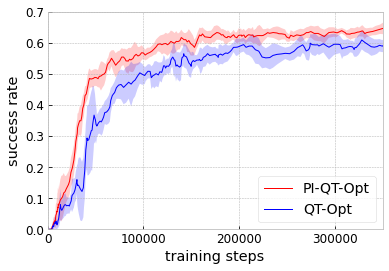}
        \caption{%
            Pick Skill
        }
    \end{subfigure}
    \begin{subfigure}{0.32\textwidth}
        \centering
        \includegraphics[width=1\linewidth]{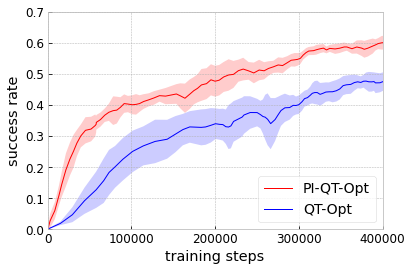}
        \caption{%
            Knock Skill
        }
    \end{subfigure}
    
    \begin{subfigure}{0.32\textwidth}
        \centering
        \includegraphics[width=1\linewidth]{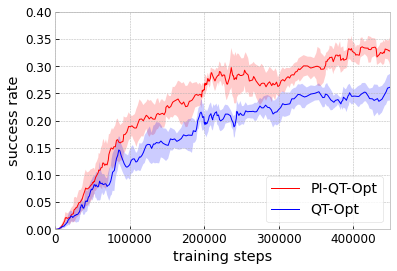}
        \caption{SayCan 297 Tasks (All Skills)}
        \label{fig:300tasks}
    \end{subfigure}
    \begin{subfigure}{0.32\textwidth}
        \centering
        \includegraphics[width=1\linewidth]{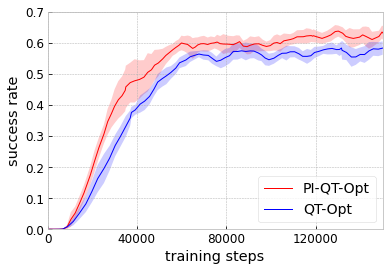}
        \caption{%
            Instance Grasping
        }
    \end{subfigure}
    \begin{subfigure}{0.32\textwidth}
        \centering
        \includegraphics[width=1\linewidth]{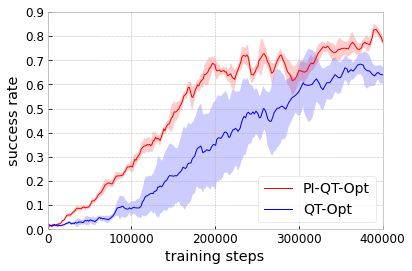}
        \caption{%
            6-Object Manipulation
        }
    \end{subfigure}
    \caption{Performance (success rate) on tasks that are used for agent training.}
    \label{fig:saycan_train}
\end{figure}

\paragraph{Performance on training tasks.}
We learn PI-QT-Opt and QT-Opt models for each of the experiment settings introduced in~\Cref{subsec:env_overview}.
\Cref{fig:saycan_train} shows evaluation results on tasks that are used for agent training in simulation. 
We can see that PI-QT-Opt consistently outperforms QT-Opt in all settings throughout training, improving the move model by 20\% and the 297-tasks model by 25\% relatively for example.
This empirically validates our hypothesis that training with the predictive information auxiliary loss leads to better and more efficient learning of general-purpose agents.

\begin{figure}[ht]
    \centering
    \begin{subfigure}{0.4\textwidth}
        \centering
        \includegraphics[width=1\linewidth]{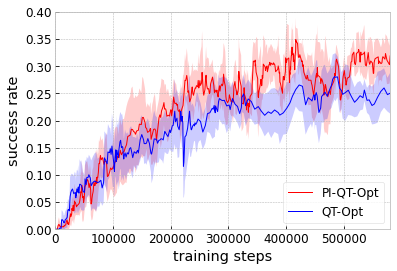}
        \caption{%
            Move Skill
        }
    \end{subfigure}
    \begin{subfigure}{0.4\textwidth}
        \centering
        \includegraphics[width=1\linewidth]{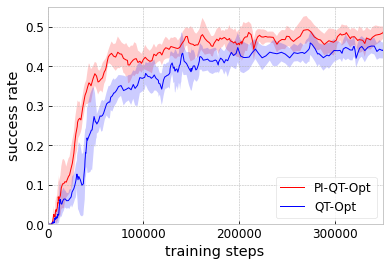}
        \caption{%
            Pick Skill
        }
    \end{subfigure}
    
    \begin{subfigure}{0.4\textwidth}
        \centering
        \includegraphics[width=1\linewidth]{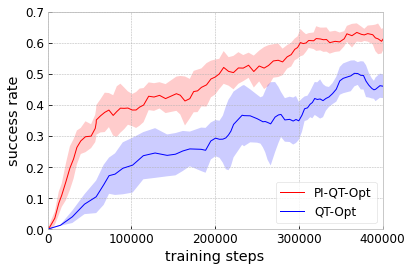}
        \caption{%
            Knock Skill
        }
    \end{subfigure}
    \begin{subfigure}{0.4\textwidth}
        \centering
        \includegraphics[width=1\linewidth]{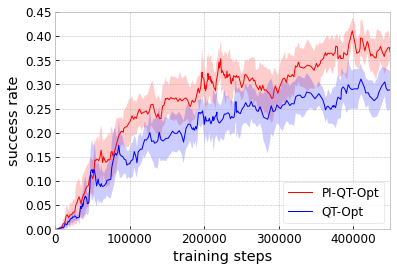}
        \caption{SayCan 297 Tasks (All Skills)}
        \label{fig:300tasks_test}
    \end{subfigure}
    \caption{Performance (success rate) on unseen novel tasks.}
    \label{fig:saycan_test}
\end{figure}

\paragraph{Zero-shot transfer to unseen tasks.}
We use the held-out SayCan tasks to evaluate zero-shot transfer of PI-QT-Opt and QT-Opt models.
A SayCan task is a composition of a skill and a set of target objects (1-2) that the robot should interact with, as described in \Cref{subsec:env_overview}. 
Here, we consider two types of zero-shot transfer:
(1) the task is never seen during training, but the target objects have been seen during training in other tasks, and (2) not only the task but the object is never seen during training. 
For (1), we evaluate the move skill models on held-out move tasks, and the 297-tasks models on held-out tasks of all skills\footnote{%
    The Move Skill training task set contain 246 tasks and each involves 2 objectives. Due to the nature of this task set, it is difficult to avoid seen objects in held-out tasks. 
    Therefore, we only use it for type (1) zero-shot transfer but not type (2).
}.
For (2), we evaluate the pick/knock skill models on held-out pick/knock tasks.
As demonstrated in \Cref{fig:saycan_test}, PI-QT-Opt outperforms QT-Opt in all the settings, showing that using the predictive information auxiliary loss leads to better zero-shot transfer to novel task compositions and unseen objects.
The 297-tasks model, for instance, is improved by 28\% relatively.

\subsection{Evaluation in the Real World}
\label{subsec:real_eval}
We directly deploy our PI-QT-Opt and QT-Opt models trained in simulation for SayCan skills on a real robot in the SayCan kitchen environment.
As described in \Cref{subsec:env_overview}, these models use image-based task context conditioning that requires information about the initial object locations (\Cref{subsec:context_conditioning}). 
Unlike the simulated environments, the ground truth locations are not available in the real environments.
Therefore, we use a VILD \citep{gu2021open} model to detect objects conditioned on object names to locate the target objects associated with the names.
When the VILD model fails to detect the target objects, we adjust the scene slightly until VILD succeeds and then continue with the evaluation.
This allows us to evaluate the policy performance exclusively.
We evaluate the SayCan Pick, Move, and Knock models; the results are presented in \Cref{tab:real_eval}, showing a clear advantage for PI-QT-Opt. 
This empirically supports our hypothesis that using the predictive information auxiliary loss enables better performance when transferring to the real world.

\begin{table}[!htb]
\begin{center}
\renewcommand{\arraystretch}{1.2}
\caption{Evaluations on the real robot (mean and standard deviation over 3 evaluations)}
\begin{tabular}{l|rrr}

\textbf{Task} & \textbf{PI-QT-Opt Success Rate} & \textbf{QT-Opt Success Rate} & \textbf{Relative Change}\\
 \hline
 \hline
SayCan Move & \textbf{22.9} $\pm$ 8.4\textbf{\%} & 13.93 $\pm$ 3.2\% & +64.4\% \\
SayCan Pick &  \textbf{42.0} $\pm$ 9.9\textbf{\%} & 28.7 $\pm$ 8.0\% & +46.6\%\\
SayCan Knock &  \textbf{54.6} $\pm$ 2.4\textbf{\%} & 36.2 $\pm$ 11.6\% & +50.7\%\\

\end{tabular}
\label{tab:real_eval}
\end{center}
\end{table}

\subsection{How does Predictive Information Relate to Agent Performance?}
\label{subsec:pi_analysis}
\begin{figure}[ht]
    \centering
    \includegraphics[width=0.5\linewidth]{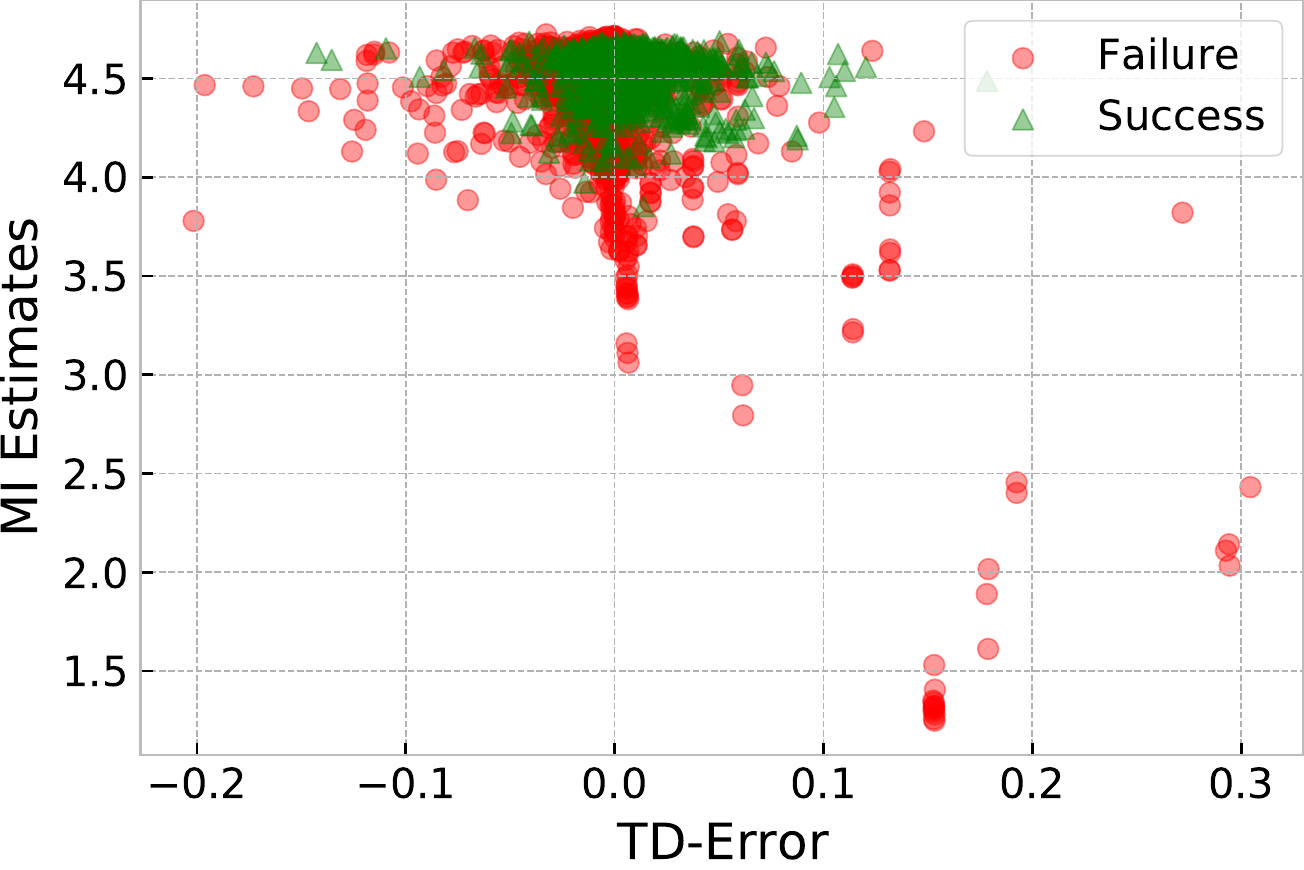}
    \caption{%
        Predictive information estimate versus TD-error.
        Each point is averaged over an episode.
        The predictive information is the mutual information (MI) between the past and future.
    }
    \label{fig:td_v_mi}
\end{figure}

A core hypothesis of this work is that the ability to model what will happen next is critical to success on control tasks.
This ability can be quantified by the amount of predictive information, $I(X,Y)$, the agent's representation captures.
In this section, we analyze the SayCan 300-task PI-QT-Opt model.
We compare the estimates of $I(X,Y)$, $\mathbb{E}[ \log b(z|y) - \log \frac{1}{K} \sum_{k=1}^K b(z|y_k)]$ (from \Cref{eq:ceb}) in successful and failed episodes versus TD-error in \Cref{fig:td_v_mi}\footnote{%
    For these predictive information analyses, we collect data for each task with a converged policy, and use a batch size of 128, which corresponds to an $I(X,Y)$ upper bound of $\log 128 = 4.852$ in order to fit these analyses into one machine, while the distributed training batch size is 4096.
}.
We can observe that the amount of predictive information is generally higher in successful episodes, and that episodes with high TD-errors have much lower predictive information and are always failures.

\section{Conclusion}
\label{sec:conclusion}

We have shown that using the predictive information auxiliary with a QT-Opt agent, i.e. PI-QT-Opt, results in faster training, higher final performance, and better generalization to unseen tasks for a single generalist agent trained on hundreds of tasks.
We have also shown that PI-QT-Opt and QT-Opt can be made to support multiple tasks by adding simple task conditioning.
Our system, PI-QT-Opt, is a generalist agent capable of solving hundreds of real-world tasks in a simple kitchen environment, in spite of having only been trained in simulation.

\paragraph{Limitations.}
We focused on a single robotic arm and gripper, so we cannot speculate on how well our approach would work on different robotic setups.
We limited our real-world experiments to simple, carefully-controlled ``kitchen'' settings, and can say nothing about performance on different types of environments.
Most importantly, our models do not have any safety guarantees, and we did not attempt to evaluate how they would perform in the presence of other agents, such as humans or animals.
Using these models in settings where there are other agents could lead to injury or death.
These limitations may be mitigated in future research by working with a range of robot platforms, expanding the breadth of tasks considered, including other agents in the environments during training and evaluation, and integrating safety systems.

\clearpage
\acknowledgments{%
The authors would like to thank 
Alex Herzog, Mohi Khansari, Daniel Kappler, Peter Pastor for adapting infrastructure and algorithms for the image-based task context from generic to instance specific grasping, 
Sangeetha Ramesh for leading robot operations for data collection and evaluations for the VILD model training,
and Kim Kleiven for leading the waste sorting service project that constitutes the framework for training and deployment of the instance grasping task set, including defining benchmark and protocol.
We thank Jornell Quiambao, Grecia Salazar, Jodilyn Peralta, Justice Carbajal, Clayton Tan, Huong T Tran, Emily Perez, Brianna Zitkovich and Jaspiar Singh for helping administrate real-world robot experiments.
We would also like to thank Sergio Guadarrama and Karol Hausman for valuable feedback.
}

\bibliography{main}  %

\clearpage
\appendix
\section{Contribution Statement}
\label{sec:contribution_statement}

\textbf{Kuang-Huei Lee:} Proposed the project initially, implemented PI-QT-Opt, experimented with SayCan settings both in sim and real, contributed to paper writing, and led the project in general.

\textbf{Ted Xiao:} Implemented simulation experiments, contributed to paper writing.

\textbf{Adrian Li:} Designed, implemented and experimented with different options of creating the target conditioning mask for instance grasping.

\textbf{Paul Wohlhart:} Designed, implemented and experimented with the neural network structure for target mask conditioned instance grasping.

\textbf{Ian Fischer:} Helped design ablation and analysis, contributed to paper writing and editing.

\textbf{Yao Lu:} Implemented PI-QT-Opt, tuned and experimented with SayCan settings both in sim and real, contributed to paper writing.

\section{Implementation Details}
\label{sec:impl_details}

\subsection{The Predictive Information Auxiliary Loss}
\label{subsec:pi_aux_details}

Here, we describe more details on the predictive information auxiliary loss introduced in \Cref{subsec:piqtopt}.
CEB~\citep{fischer2020conditional} and InfoNCE~\citep{poole2019variational} require two encoder distributions $e(z|x)$ and $b(z|y)$.
\citet{fischer2020conditional} defines $e(z|x)$ to be the \emph{forward encoder} from which the representation $z$ is sampled and $b(z|y)$ to be a variational \emph{backward encoder} that approximates the unknown density $p(z|y) = \int dx\, \frac{p(x,y,z)}{p(y)}$.
In this work, we follow~\citep{lee2021compressive} to choose $e(z|x)$ and $b(z|y)$ to be parameterized by von Mises-Fisher (vMF) distributions, which empirically yields good performance in learning self-supervised visual representations.

The von Mises-Fisher is a distribution on the $(n-1)$-dimensional hypersphere. 
The probability density function is given by $f_n(z, \mu, \kappa) = C_n(\kappa)\exp(\kappa \mu^Tz)$, where $\mu$ and $\kappa$ are the mean direction and concentration parameter respectively.
We assume $\kappa$ is a constant.
The normalization term $C_n(\kappa)$ is a function of $\kappa$ and equal to $\frac{\kappa^{n/2-1}}{(2\pi)^{n/2} I_{n/2-1}(\kappa)}$, where $I_v$ denotes the modified Bessel function of the first kind at order $v$.

As shown in~\citep{lee2021compressive}, when the forward encoder concentration parameter $\kappa_e$ approaches infinity, $\kappa_e \to \infty$, $e(z|x)$ becomes a spherical delta distribution and InfoNCE that uses vMF distributions reduces to the commonly used deterministic form of InfoNCE with cosine similarity as its distance function.

In our implementation, we parameterize $e(z|x)$ and $b(z|y)$ as follows. 
We select $\kappa_e=8192$ and $\kappa_b=7$. 
$\mu_e$ is a $64$d vector coming from an MLP with two $512$d hidden layers on top of a convolution network shared with the online Q-network.
Similarly, $\mu_b$ is a $64$d vector coming from an MLP with two $512$d hidden layers on top of a convolution network shared with the lagged target Q-network, which is not updated using gradients.
$\beta$ in \Cref{eq:ceb} is a Lagrange multiplier that controls the strength of information compression.
We choose $\beta=0.01$.

The predictive information auxiliary objective (\Cref{eq:ceb}) and the Q-learning loss (\Cref{eq:bellman_loss}) are weighted combined and learned with the same optimizer, with 1.0 on the Q-learning loss and 0.01 on the CEB loss.

\subsection{Implementation Details of Multi-Task Context Conditioning}
\label{sec:task_conditioning_details}

We introduced two implementations of task context conditioning for multi-task learning in \Cref{subsec:context_conditioning}: image-based and language-based.
In this section, we will dive into more details about these two implementations.

\paragraph{Image-based Task Context Conditioning.}
Our image-based implementation is similar to \citet{bai2017mask}, which utilizes image segmentation masks for task conditioning.
In our setting, a manipulation task involves a robot skill (e.g. move, pick, knock) and a set of objects of interest.
For any given task, we localize each object of interest with a 10x10 colored square mask in an overlay image, where the color determines the semantic skill pertaining to that object.
Each square mask is centered at the the object bounding box center and agnostic of the object size.
For example, the blue square in the first frame of \Cref{fig:experiments}(f) indicates that the task is ``picking up the can on the right'', and the green square in the first frame of \Cref{fig:experiments}(g) represents a task ``knock down the plastic bottle in the middle''.
Note that the RGB color is for better presentation, and the actual implementation represents tasks in grayscale.

The reason that we use this type of task context mask instead of pixel-accurate segmentation masks~\citep{bai2017mask} is as follows.
While perfect pixel-accurate masks can be easily provided in simulation, when deployed in the real world (\Cref{subsec:real_eval}), pixel-accurate mask boundaries can be sensitive to many conditions including lighting, occlusions, the angle of view, and other perturbations.
In practice, we find that simple 10x10 square masks tend to be more robust to these types of noise.

Notably, we only produce the task context masks one time per episode at the first frame.
The same overlay image and initial frame are used as the task context throughout the entire episode. 
They are used only to represent a task, and not for enhancing perception during planning.
This also avoids the need to run the VILD model~\citep{gu2021open} for object detection in real-time when deployed on the real robot.

\paragraph{Language-based Context Conditioning.}
Language Contexts are computed by using a pretrained and frozen Universal Sentence Encoder (USE)~\cite{cer2018universal} to embed natural language task instructions, where each task corresponds to exactly one natural language instruction.

\begin{figure}[t]
    \centering
    \includegraphics[width=1.0\linewidth]{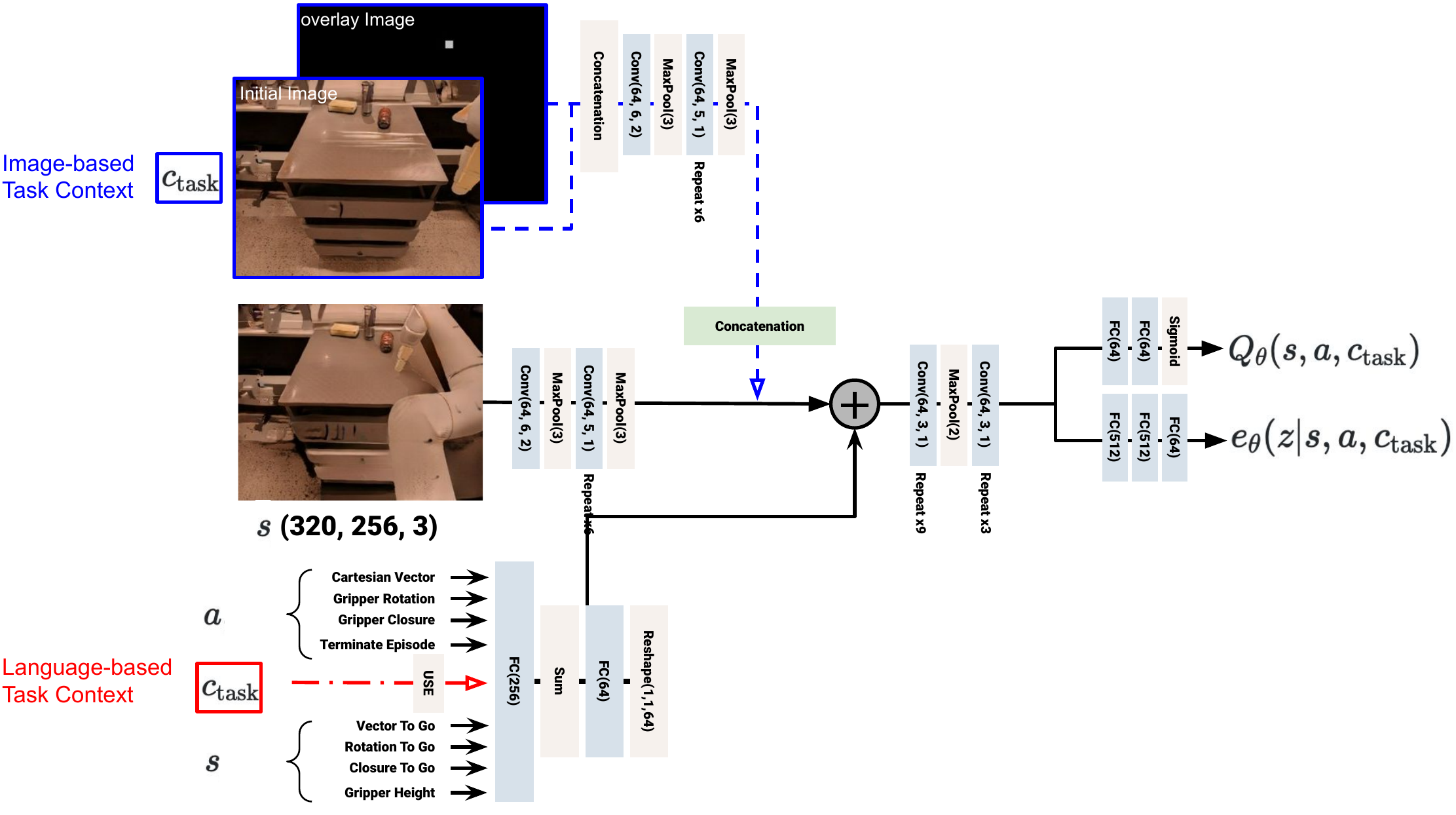}
    \caption{
        Network Architecture. 
        Task context are either image-based (blue path) or language-based (red path).
        The convolution parameters shown in this diagram are (channels, kernel size, stride size).
    }
    \label{fig:architecture}
\end{figure}

\subsection{Architecture}
\Cref{fig:architecture} shows the detailed network architecture. 
The convolutional and MLP blocks are similar to the network architecture used in \cite{kalashnikov2018scalable}. 
Specifically, the first convolutional block before merging with the action or task context contains six convolutional layers and two pooling layers.
The second convolutional block after merging contains nine convolutional layers.
The Q-value MLP block contains two dense layers.
Each convolutional and hidden dense layer is followed by a batch norm layer~\citep{ioffe2015batch} and a ReLU activation layer.

An image-based task context consists of two images: one is the initial RGB image and the other is the corresponding grayscale overlay image.
We concatenate these two images along the channel dimension and apply a convolutional block that has the same architecture as the first convolutional block that processes the current observation image.
We concatenate the image-based task context embedding with the current observation image embedding along the channel dimension.
Following this step, the representation vector of action and proprioceptive state is merged with visual features by broadcasted element-wise addition. 

In the alternative setting where we use the language-based task context instead of image-based task context, we use the natural language Universal Sentence Encoder (USE)~\cite{cer2018universal} embedding of the current task instruction.
This USE embedding is fed into an MLP with two fully connected layers, each with a batch norm layer.
Finally, the processed embedding is fused with the action and proprioceptive state and eventually merged with visual features.

\subsection{Comparing PI-QT-Opt and PI-SAC}
\label{sec:qtopt_v_sac}
We chose QT-Opt as the underlying control algorithm in this work.
Compared to SAC~\cite{haarnoja2018soft} that PI-SAC~\cite{lee2020predictive} used, the main advantage of QT-Opt is that the action selection is pure sampling-based (CEM) \cite{rubinstein2004cross}, and thus does not require a gradient-learned actor as in SAC. 
This makes it possible to have complex and even dynamic action space and bounds without worrying about how to back-propagate gradients.

This makes adding safety constraints simple. 
Every time when we sample an action, we can clip the action according to the action bound, which can change based on the safety constraints at each specific robot state. 
For a gradient-based actor, such clipping zero-outs gradients, making optimization challenging. 
We did try training SAC and it did not learn with safety-constraint action clipping. 
How to make a gradient-based actor work in our setting is still an open-ended research question, for which we do not have a good answer at the moment.

On the other hand, there are certainly room for improvements of the underlying control algorithm in aspects that we do not attempt to address explicitly in this work, such as exploration.

There are a few other architecture differences between PI-QT-Opt and PI-SAC. 
Because PI-SAC was evaluated on DM-Control~\cite{tassa2018dmcontrol}, it follows the standard approach to stack 3 frames and, in order to avoid overlapping between the past and the future frames.
The backward encoder does not share its convolutional backbone with the target Q-network.
In PI-QT-Opt, because we only consider 1 past frame and 1 future frame, we are able to simplify the design, making backward encoder share the convolutional backbone with the target Q-network.
In PI-SAC, the forward and backward encoder distributions are parameterized as Gaussian distributions, whereas, in PI-QT-Opt, the forward and backward encoder distributions are parameterized as von Mises-Fisher distributions.

\subsection{Concurrent Control and Blocking Control}
\label{subsec:concurrent_control_details}
\begin{figure}[!t]
    \centering
    \begin{subfigure}{0.45\textwidth}
        \centering
        \includegraphics[width=1\linewidth]{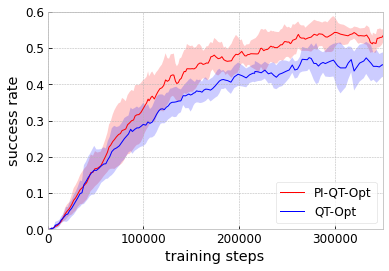}
        \caption{%
            On tasks that are used for agent training.
        }
    \end{subfigure}
    \begin{subfigure}{0.45\textwidth}
        \centering
        \includegraphics[width=1\linewidth]{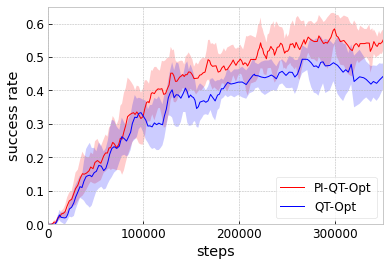}
        \caption{%
            On unseen novel tasks.
        }
    \end{subfigure}
    \caption{Performance on SayCan Move tasks (Blocking Control)}
    \label{fig:blocking_ablation}
\end{figure}

As described in \Cref{subsec:env_overview}, we utilize blocking control in the simulation-only environments (Instance Grasping and 6-Object Manipulation), where the policy waits until the previous action completes before observing the next environment state and planning the next action.
Motivated by faster and more reactive robot motions for real world evaluations, we utilize concurrent control~\citep{xiao2020nonblocking} in all SayCan experiments, which means that the current action is computed while the previous action is still executing.
In particular, the SayCan results reported in \Cref{fig:saycan_train}, \Cref{fig:saycan_test}, \Cref{tab:real_eval}, and \Cref{tab:real_eval_2} all utilize concurrent control.
Intuitively, concurrent control is more challenging than blocking control, since predicting actions to execute while the robot is moving involves implicit planning compared to blocking control, where actions are guaranteed to execute in the exact same state as the observation.
In Figure~\ref{fig:blocking_ablation}, we present PI-QT-Opt and QT-Opt results on a blocking version of the SayCan Move task set as an example.
Compared with \Cref{fig:saycan_train}(a) and \Cref{fig:saycan_test}(a), we can observe that models learn faster and achieve better final performance when utilizing blocking control.
The results also show that PI-QT-Opt outperforms QT-Opt by a similar amount on both the blocking and continuous control versions of the tasks (compare \Cref{fig:saycan_train}(a) and \Cref{fig:saycan_test}(a) to \Cref{fig:blocking_ablation}(a) and (b)).

\subsection{Model Training Details}
\label{subsec:training_details}
We base our implementation on the distributed asynchronous QT-Opt system introduced in~\citet{kalashnikov2018scalable}.
Our system uses the TF-Agents RL library~\cite{TFAgents}.
For each experiment, we use 3000 data collection workers to interact with the simulation environment, 3000 ``Bellman updater`` jobs (Section 4.3 of \cite{kalashnikov2018scalable}), a distributed replay buffer spread over 20 workers (Section F.5 of \cite{kalashnikov2018scalable}), and 16 TPUv2 for model learning. 
We learn with stochastic gradient descent (SGD) with momentum.
The learning rate is $9.56\times10^{-3}$ and the momentum weight is $0.984$.
The model training time depends on the learning task set.
For example, a SayCan 297-task model takes five to seven days to learn, while an instance grasping model takes 20 hours.

\subsection{Model Evaluation Protocol Details}
\label{subsec:eval_details}
In simulation, for every episode, we sample a task, which involves a skill and objects of interest.
The objects of interest and additional randomly sampled distractor objects are then randomly placed in the scene.
If the randomly generated scene is already in a successful state, we regenerate it. 
The robot is always initialized with the same pose of the arm but the base position is randomly sampled within a small rectangular area in front of the counter (for SayCan tasks), the waste station (for Instance Grasping tasks), or the table (for 6-Object Manipulation tasks).

For controlled and fair evaluations in the real-world tasks, we use up to three variations of each task that are manually reset as precisely as possible for each model we evaluate (three models each for QT-Opt and PI-QT-Opt).
Thus, each model gets a single attempt at each task variation, but multiple attempts at each task, and each model sees the same set of task variations.
The only variation between models in evaluation is randomization of the robot's base position at the start of each episode.

\section{SayCan Objects}
\label{sec:saycan_obj}

\begin{figure}[ht]
    \centering
    \includegraphics[width=1\linewidth]{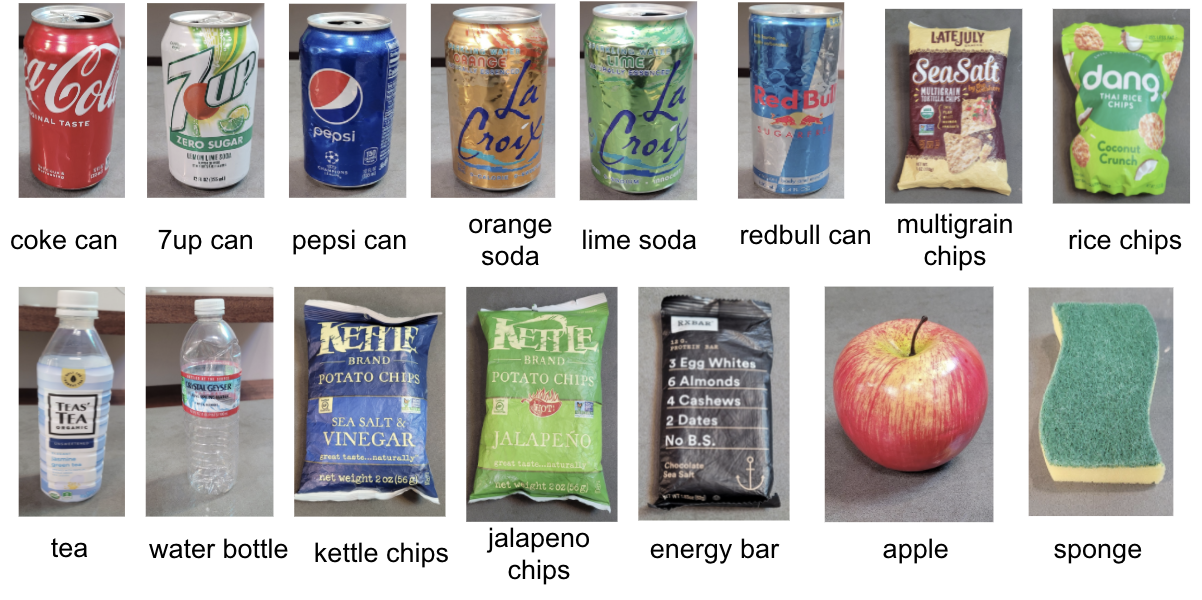}
    \caption{%
        Example objects used in the SayCan tasks. 
        Objects are placed on a kitchen counter, where a robot must perform different manipulation skills such as picking objects up, moving objects, and knocking objects over. 
        We train on 17 objects in simulation and evaluate on these 15 objects in the real world.
        Not shown here are a blue energy bar and an orange.
    }
    \label{fig:saycan_objects}
\end{figure}

\Cref{fig:saycan_objects} shows examples of the objects used in the SayCan tasks.

\section{Evaluating the SayCan-297-Task Model in the Real World}

In \Cref{subsec:real_eval}, we evaluated per-skill SayCan models in the real world.
Here, we additionally report real-world evaluation of the SayCan-297-Task model, which is trained on all 297 SayCan tasks.
The evaluation results on each skill category are summarized in \Cref{tab:real_eval_2}.
In this set of experiments, PI-QT-Opt continues to outperform QT-Opt by a large margin. 

\begin{table}[!t]
\begin{center}
\renewcommand{\arraystretch}{1.2}
\caption{Evaluations of a single model that solves SayCan 297 Tasks on the real robot.}
\begin{tabular}{l|rrr}

\textbf{Task} & \textbf{PI-QT-Opt Success Rate} & \textbf{QT-Opt Success Rate} & \textbf{Relative Change}\\
 \hline
 \hline
SayCan Move & \textbf{25.0} $\pm$ 3.2\textbf{\%} & 17.4 $\pm$ 4.3\% & +44.0\%\\
SayCan Pick & \textbf{33.4} $\pm$ 12.5\textbf{\%} & 22.2 $\pm$ 9.6\% & +50.1\%\\
SayCan Knock & \textbf{52.3} $\pm$ 10.7\textbf{\%} & 39.2 $\pm$ 4.1\% & +33.2\%\\

\end{tabular}
\label{tab:real_eval_2}
\end{center}
\end{table}

\section{Ablation of Predictive Information Compression}
\label{sec:compression_exp}
\begin{figure}[htb]
    \centering
    \begin{subfigure}{0.49\textwidth}
        \centering
        \includegraphics[width=1\linewidth]{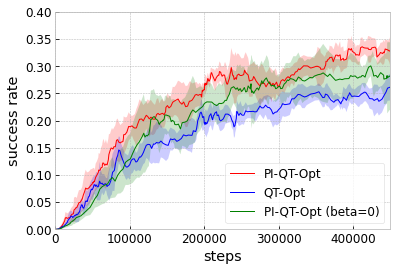}
        \caption{%
            SayCan 297 Tasks (evaluated on training tasks).
        }
    \end{subfigure}
    \hfill
    \begin{subfigure}{0.49\textwidth}
        \centering
        \includegraphics[width=1\linewidth]{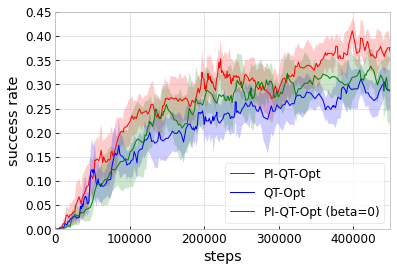}
        \caption{%
            SayCan 297 tasks (evaluated on unseen tasks).
        }
    \end{subfigure}
    \caption{%
        Comparison between PI-QT-Opt ($\beta=0.01$ by default), PI-QT-Opt ($\beta=0.0$, no explicit compression), and QT-Opt.
    }
    \label{fig:moveskill}
\end{figure}

To understand the importance of compression to PI-QT-Opt, rather than just predicting the future, we compare QT-Opt, PI-QT-Opt at $\beta=0$, and PI-QT-Opt at $\beta=0.01$, the value used in all other experiments.
When $\beta=0$, the model still learns to predict the future ($Y$) from the past ($X$), but it no longer makes any explicit attempt to compress irrelevant information in $X$.
In \Cref{fig:moveskill} we compare these three models on the SayCan 297 Tasks, both for training tasks (a), and unseen evaluation tasks (b).
In both cases, the compressed version of PI-QT-Opt slightly outperforms the uncompressed version, which still slightly outperforms QT-Opt without the predictive information auxiliary loss.
We note, however, that compressed PI-QT-Opt often overlaps in performance with uncompressed PI-QT-Opt, so clearly the advantage of PI-QT-Opt over QT-Opt is only partially due to compression.

\section{Extending the Image-Based Task Context to Free-form Commands}
\label{sec:extend_nat_lang}

The tasks we selected are in a structured language form of skill and target object sets.
To support more complicated, natural language commands, we could encode the task description with a language model.
However, for the image-based task context described in \Cref{subsec:context_conditioning} and \Cref{sec:task_conditioning_details}, we would also need a mechanism to visually represent the command's target(s) in the scene.
We speculate that providing object detection network activations, rather than the simpler conditioning described in \Cref{subsec:context_conditioning}, may be sufficient to capture the relevant information for interpreting the command.
We leave this generalization of our method for future work.

\newpage
\section{Lists of 297 SayCan Tasks}
\label{sec:saycan_task_list}

In the following table we list all 297 SayCan tasks that are used for training or held-out for evaluation.
Note that the knock skill target objects only include cans and bottles since other objects cannot be ``knocked down`` from an upright pose (See~\Cref{sec:saycan_obj} for the object list).

\begin{table}[!ht]
\scriptsize
\begin{tabular}{ll}
\toprule
\bf{Pick skill} & \\
\hline
\bf{Training tasks} & \\
pick 7up can &
pick apple \\
pick blue chip bag &
pick brown chip bag \\
pick coke can &
pick green can \\
pick green jalapeno chip bag &
pick orange can \\
pick pepsi can &
pick redbull can \\
pick rxbar blueberry &
pick water bottle \\
\bf{Held-out evaluation tasks} & \\
pick blue plastic bottle &
pick green rice chip bag \\
pick orange &
pick rxbar chocolate \\
pick sponge \\
\hline
\bf{Knock skill} & \\
\hline
\bf{Training tasks} & \\
knock 7up can over &
knock blue plastic bottle over \\
knock coke can over &
knock green can over \\
knock pepsi can over &
knock redbull can over \\
knock water bottle over \\
\bf{Held-out evaluation tasks} & \\
knock orange can over \\
\hline
\bf{Move skill} & \\
\hline
\bf{Training tasks} & \\
move 7up can near apple &
move 7up can near blue chip bag \\
move 7up can near blue plastic bottle &
move 7up can near brown chip bag \\
move 7up can near coke can &
move 7up can near green can \\
move 7up can near green jalapeno chip bag &
move 7up can near green rice chip bag \\
move 7up can near orange &
move 7up can near orange can \\
move 7up can near pepsi can &
move 7up can near rxbar blueberry \\
move 7up can near rxbar chocolate &
move 7up can near sponge \\
move 7up can near water bottle &
move apple near 7up can \\
move apple near blue chip bag &
move apple near blue plastic bottle \\
move apple near brown chip bag &
move apple near coke can \\
move apple near green can &
move apple near green jalapeno chip bag \\
move apple near orange &
move apple near orange can \\
move apple near pepsi can &
move apple near rxbar blueberry \\
move apple near rxbar chocolate &
move apple near sponge \\
move apple near water bottle &
move blue chip bag near 7up can \\
move blue chip bag near apple &
move blue chip bag near brown chip bag \\
move blue chip bag near coke can &
move blue chip bag near green can \\
move blue chip bag near green jalapeno chip bag &
move blue chip bag near green rice chip bag \\
move blue chip bag near orange &
move blue chip bag near orange can \\
move blue chip bag near redbull can &
move blue chip bag near rxbar blueberry \\
move blue chip bag near rxbar chocolate &
move blue chip bag near water bottle \\
move blue plastic bottle near 7up can &
move blue plastic bottle near apple \\
move blue plastic bottle near blue chip bag &
move blue plastic bottle near brown chip bag \\
move blue plastic bottle near coke can &
move blue plastic bottle near green can \\
move blue plastic bottle near green jalapeno chip bag &
move blue plastic bottle near green rice chip bag \\
move blue plastic bottle near orange &
move blue plastic bottle near orange can \\
move blue plastic bottle near pepsi can &
move blue plastic bottle near redbull can \\
move blue plastic bottle near rxbar blueberry &
move blue plastic bottle near rxbar chocolate \\
move blue plastic bottle near sponge &
move blue plastic bottle near water bottle \\
move brown chip bag near 7up can &
move brown chip bag near blue chip bag \\
move brown chip bag near blue plastic bottle &
move brown chip bag near coke can \\
move brown chip bag near green can &
move brown chip bag near green jalapeno chip bag \\
move brown chip bag near orange &
move brown chip bag near orange can \\
move brown chip bag near pepsi can &
move brown chip bag near rxbar blueberry \\
move brown chip bag near rxbar chocolate &
move brown chip bag near sponge \\
move brown chip bag near water bottle &
move coke can near 7up can \\
move coke can near apple &
move coke can near blue chip bag \\
move coke can near blue plastic bottle &
move coke can near brown chip bag \\
move coke can near green can &
move coke can near green rice chip bag \\
move coke can near orange &
move coke can near orange can \\
move coke can near pepsi can &
move coke can near redbull can \\
move coke can near rxbar blueberry &
move coke can near rxbar chocolate \\
move coke can near sponge &
move green can near blue chip bag \\
move green can near blue plastic bottle &
move green can near brown chip bag \\
move green can near green jalapeno chip bag &
move green can near green rice chip bag \\
move green can near orange &
move green can near orange can \\
\end{tabular}
\end{table}

\clearpage
\begin{table}[!ht]
\scriptsize
\begin{tabular}{ll}
move green can near pepsi can &
move green can near redbull can \\
move green can near rxbar blueberry &
move green can near rxbar chocolate \\
move green can near sponge &
move green can near water bottle \\
move green jalapeno chip bag near 7up can &
move green jalapeno chip bag near apple \\
move green jalapeno chip bag near blue plastic bottle &
move green jalapeno chip bag near brown chip bag \\
move green jalapeno chip bag near coke can &
move green jalapeno chip bag near green can \\
move green jalapeno chip bag near green rice chip bag &
move green jalapeno chip bag near orange \\
move green jalapeno chip bag near orange can &
move green jalapeno chip bag near pepsi can \\
move green jalapeno chip bag near redbull can &
move green jalapeno chip bag near rxbar blueberry \\
move green jalapeno chip bag near rxbar chocolate &
move green jalapeno chip bag near sponge \\
move green jalapeno chip bag near water bottle &
move green rice chip bag near 7up can \\
move green rice chip bag near apple &
move green rice chip bag near blue chip bag \\
move green rice chip bag near blue plastic bottle &
move green rice chip bag near brown chip bag \\
move green rice chip bag near coke can &
move green rice chip bag near green can \\
move green rice chip bag near green jalapeno chip bag &
move green rice chip bag near pepsi can \\
move green rice chip bag near redbull can &
move green rice chip bag near rxbar blueberry \\
move green rice chip bag near rxbar chocolate &
move green rice chip bag near sponge \\
move green rice chip bag near water bottle &
move orange can near 7up can \\
move orange can near apple &
move orange can near blue chip bag \\
move orange can near blue plastic bottle &
move orange can near coke can \\
move orange can near green can &
move orange can near green jalapeno chip bag \\
move orange can near green rice chip bag &
move orange can near orange \\
move orange can near pepsi can &
move orange can near redbull can \\
move orange can near rxbar blueberry &
move orange can near rxbar chocolate \\
move orange can near sponge &
move orange can near water bottle \\
move orange near 7up can &
move orange near apple \\
move orange near blue chip bag &
move orange near blue plastic bottle \\
move orange near brown chip bag &
move orange near coke can \\
move orange near green can &
move orange near green jalapeno chip bag \\
move orange near green rice chip bag &
move orange near orange can \\
move orange near pepsi can &
move orange near redbull can \\
move orange near rxbar blueberry &
move orange near rxbar chocolate \\
move orange near sponge &
move orange near water bottle \\
move pepsi can near 7up can &
move pepsi can near apple \\
move pepsi can near blue chip bag &
move pepsi can near blue plastic bottle \\
move pepsi can near brown chip bag &
move pepsi can near coke can \\
move pepsi can near green can &
move pepsi can near green jalapeno chip bag \\
move pepsi can near green rice chip bag &
move pepsi can near orange \\
move pepsi can near redbull can &
move pepsi can near rxbar blueberry \\
move pepsi can near rxbar chocolate &
move pepsi can near sponge \\
move pepsi can near water bottle &
move redbull can near 7up can \\
move redbull can near apple &
move redbull can near blue chip bag \\
move redbull can near blue plastic bottle &
move redbull can near brown chip bag \\
move redbull can near green can &
move redbull can near green jalapeno chip bag \\
move redbull can near green rice chip bag &
move redbull can near orange \\
move redbull can near orange can &
move redbull can near pepsi can \\
move redbull can near rxbar blueberry &
move redbull can near rxbar chocolate \\
move redbull can near sponge &
move redbull can near water bottle \\
move rxbar blueberry near 7up can &
move rxbar blueberry near apple \\
move rxbar blueberry near blue chip bag &
move rxbar blueberry near brown chip bag \\
move rxbar blueberry near coke can &
move rxbar blueberry near green can \\
move rxbar blueberry near green jalapeno chip bag &
move rxbar blueberry near green rice chip bag \\
move rxbar blueberry near orange &
move rxbar blueberry near pepsi can \\
move rxbar blueberry near redbull can &
move rxbar blueberry near rxbar chocolate \\
move rxbar blueberry near sponge &
move rxbar blueberry near water bottle \\
move rxbar chocolate near 7up can &
move rxbar chocolate near apple \\
move rxbar chocolate near blue chip bag &
move rxbar chocolate near blue plastic bottle \\
move rxbar chocolate near brown chip bag &
move rxbar chocolate near coke can \\
move rxbar chocolate near green can &
move rxbar chocolate near green jalapeno chip bag \\
move rxbar chocolate near green rice chip bag &
move rxbar chocolate near orange \\
move rxbar chocolate near orange can &
move rxbar chocolate near pepsi can \\
move rxbar chocolate near redbull can &
move rxbar chocolate near sponge \\
move rxbar chocolate near water bottle &
move sponge near 7up can \\
move sponge near blue chip bag &
move sponge near blue plastic bottle \\
move sponge near brown chip bag &
move sponge near coke can \\
move sponge near green can &
move sponge near green jalapeno chip bag \\
move sponge near green rice chip bag &
move sponge near orange \\
move sponge near orange can &
move sponge near pepsi can \\
move sponge near redbull can &
move sponge near rxbar blueberry \\
move sponge near rxbar chocolate &
move sponge near water bottle \\
move water bottle near apple &
move water bottle near blue chip bag \\
move water bottle near blue plastic bottle &
move water bottle near brown chip bag \\
move water bottle near coke can &
move water bottle near green can \\
move water bottle near green jalapeno chip bag &
move water bottle near green rice chip bag \\
move water bottle near orange &
move water bottle near orange can \\
move water bottle near pepsi can &
move water bottle near redbull can \\
move water bottle near rxbar blueberry &
move water bottle near rxbar chocolate \\
\end{tabular}
\end{table}

\clearpage
\begin{table}[ht!]
\scriptsize
\begin{tabular}{ll}
\bf{Held-out evaluation tasks} & \\
move 7up can near redbull can &
move apple near green rice chip bag \\
move apple near redbull can &
move blue chip bag near blue plastic bottle \\
move blue chip bag near pepsi can &
move blue chip bag near sponge \\
move brown chip bag near apple &
move brown chip bag near green rice chip bag \\
move brown chip bag near redbull can &
move coke can near green jalapeno chip bag \\
move coke can near water bottle &
move green can near 7up can \\
move green can near apple &
move green can near coke can \\
move green jalapeno chip bag near blue chip bag &
move green rice chip bag near orange \\
move green rice chip bag near orange can &
move orange can near brown chip bag \\
move pepsi can near orange can &
move redbull can near coke can \\
move rxbar blueberry near blue plastic bottle &
move rxbar blueberry near orange can \\
move rxbar chocolate near rxbar blueberry &
move sponge near apple \\
move water bottle near 7up can &
move water bottle near sponge \\
\bottomrule
\end{tabular}
\end{table}

\end{document}